\documentclass[11pt,a4paper]{article}
\usepackage[hyperref]{emnlp-ijcnlp-2019}
\usepackage{times}
\usepackage{latexsym}
\usepackage{booktabs}

\usepackage{url}
\usepackage{graphicx}

\aclfinalcopy 

\usepackage{xspace}

\title{Financial Event Extraction Using Wikipedia-Based Weak Supervision}
\author{\textbf{Liat Ein-Dor, Ariel Gera, Orith Toledo-Ronen, Alon Halfon, Benjamin Sznajder,} 
\\
\textbf{Lena Dankin, Yonatan Bilu, Yoav Katz and Noam Slonim}\\
  IBM Research, Haifa, Israel}

\date{}

\begin{document}
\maketitle

\begin{abstract}
Extraction of financial and economic events from text has previously been done mostly using rule-based methods, with more recent works employing machine learning techniques. This work is in line with this latter approach, 
leveraging relevant Wikipedia sections to extract weak labels for sentences describing economic events. Whereas previous weakly supervised approaches required a knowledge-base of such events, or corresponding financial figures, our approach requires no such additional data, and can be employed to extract economic events related to companies which are not even mentioned in the training data.
\end{abstract}

\section{Introduction}
\textit{Event Extraction} from text \cite{hogenboom2011overview, ritter2012open, hogenboom2016survey} has been the subject of active research for over two decades \cite{allan2003topic}. 
Detection and extraction of finance-related  
events have mostly focused on events described in news articles, which are likely to impact stock prices. In particular, previous work has sought to extract descriptions of events pertaining to a specific company, and analyzed how such events correlate with measures of that company's stock (price, volatility etc.).
While much of the literature has focused on the prediction of stock prices (e.g., \citealp{Ding:2015:DLE:2832415.2832572, xie2013semantic}), it is recognized that predicting future stock movements is a formidable challenge (see e.g. \citealp{merello2018}); still, there are 
use-cases that might benefit from business-related event extraction from news. 

One promising direction 
is enhancing the finance-related research performed by finance analysts. Such research typically requires reviewing a large body of news data under severe time constraints. 
We propose an automatic system for highlighting meaningful company-related news events 
that are likely to deserve the analyst's attention. 

Work on economic event extraction often defines an ad-hoc taxonomy of events, and what constitutes an 
'important event' 
for one might not be considered as such for another. For instance, the CoProE event ontology \cite{kakkonen2011} includes events such as patent issuance and delayed filing of company reports, which are not considered 
by \citet{du2016puls}; 
similarly, 
while CoProE consider 
earnings estimates by analysts as events, \citet{jacobs2018economic} examine 
instead analyst buy ratings and recommendations.

Outlining a
comprehensive list of event types seems futile. 
For example, if a company's databases are hacked, this is certainly an influential event; 
but compiling an explicit and exhaustive event taxonomy that is sufficiently fine-grained to include all events such as this one
is doomed to fail.
At the same time, a formal event hierarchy 
is not necessarily required  
from an analyst's perspective. The strength of an automated system comes from the ability to process a large volume of news data and detect events of interest; automatically classifying these events into types is 
probably of secondary importance 
to an expert in the field.

Thus, our focus here is on a binary classification problem that is not type-based. This presents an interesting challenge, since the aim is not capturing the characteristics of predefined event types, but rather capturing general properties of relevant events.

The common 
NLP approach for economic event extraction has mostly made use of hand-crafted rules and patterns \cite{feldman2011stock, Arendarenko2012, xie2013semantic, hogenboom2013semantics, ding-etal-2014-using, Ding:2015:DLE:2832415.2832572, du2016puls}. 
However, creating and maintaining such rules is time consuming, and further seems less suitable for our scenario, where no set of underlying event types (which give rise to such rules) is assumed. Hence, here we follow a different, more flexible approach, 
that relies on a robust statistical learning framework 
for identifying relevant events. 
In particular, 
we adopt a supervised learning approach for identifying events related to a given company, and suggest to train a sentence-level classifier for this purpose. Given sentences from news articles discussing the company, the 
classifier aims to identify 
sentences containing events that would be of interest to the analyst.  events. 
Since the sentences come from articles discussing the company,
our main focus is on determining whether a sentence conveys an event worth considering, %
and not on ascertaining that it is related to the company.    

Learning a supervised model requires annotated data. The standard approach for obtaining annotated data involves human annotation,
which requires a substantial effort and limits the 
size of the data, which in turn may 
hinder the results. One way to overcome this problem is using weak supervision \cite{weak_supervision}, where labelled data is generated automatically using heuristics rather than manual annotation. Although such data may be noisier and less precise compared to standard labelled data, it enables to create much larger amounts of data at a significantly lower cost. Here we rely on content from Wikipedia to automatically generate a weakly-labelled sentence dataset for company events.
We report experimental results that demonstrate the potential merit of our approach. 

\section{Related Work}

\citet{Arendarenko2012} relied on a collection of hand-crafted detection rules in order to recognize 41 distinct company-related event types, and \citet{du2016puls} used about 600 distinct patterns to cover 15 business event types.

More recently, machine-learning techniques were considered for this task. \citet{jacobs2018economic} frame the problem as a multi-class classification task. They define a taxonomy of 10 event types, in addition to a "no-event" class, and 7 companies of interest, and rely on manual annotation to train a sentence-level multi-class classifier. Testing several classifiers, they show that a linear SVM classifier attains the best results for most event types. While the current paper also adopts a supervised learning sentence-level approach, here the data is constructed based on weak labels, and the task is framed as a type-independent binary classification problem. 

\citet{ronnqvist2017bank} used weak supervision in the context of financial events, focusing on bank distress events. They consider 101 banks for which 243 such events, and their date, are known. They then extract 386K sentences referring to these banks, and consider a sentence as describing a distress event if there is a matching event in the knowledge base mentioning the same bank and occurring near the publication date of the article from which the sentence was extracted.
This approach 
requires a large knowledge-base of specific events, which is not readily available when moving from a confined event type (i.e. bank distress) to a diverse space of events. In this work we suggest a weak-label approach that aims to encompass a variety of relevant entities, event types and event occurrences.

\section{Data}
\label{data}

We used two types of datasets, one which is created automatically based on weak labels, and another which is based on manual annotation.

\subsection{Weakly labelled datasets - Wikipedia}

We leverage the content of Wikipedia articles describing 
companies as a source of influential events in the company's chronology. 

In order to automatically identify 'positive' 
sentences which likely describe noteworthy 
events, we rely on two observations: 1. Such events 
tend to appear within specific Wikipedia sections. 2. Sentences beginning with a date, specifically the \textit{date-pattern} $['On/In/By/As$ $of' + month + year]$, often describe an event. 
Thus, we manually created a lexicon of words which tend to appear in the titles of event-prone sections. 
A section whose title contains one of the following  words is defined as an \textit{event-section}: 
history, creation, leadership, corporate, acquisitions, growth, finance, financial, lawsuits, litigation, legal.

Given a company $C$, we select from its Wikipedia article all sentences appearing in an \textit{event-section} and starting with a \textit{date-pattern}. We remove the opening date and mark the sentences as positive examples with respect to $C$.
All sentences which do not start with a \textit{date-pattern} and are not in an \textit{event-section} are considered as negative. To balance the dataset, we enforce  
an equal number of positive and negative examples by discarding sentences from the larger set. In addition, since many positive examples begin with either the company's name or the words "the company", we aim to balance the two classes in terms of sentences containing these patterns. The rest of the negative examples are chosen at random.

The procedure described above was used to create two datasets. The first, $S\&P$-$wiki$, is generated from Wikipedia articles of the companies on the S\&P-500 index. A larger dataset, $Extended$-$wiki$, was later generated from Wikipedia articles of companies traded in one of five major stock exchanges\footnote{Hong Kong, London, NASDAQ, NYSE and Tokyo; Extracted via Wikipedia categories of these exchanges.}
, yielding 3.8K companies in total.

Each dataset was split into train 
and  
test sets based on dates - all positive examples up to 2018 are in the training set, and all those from 2019 are in the test set. Negative examples, which have no date attached, were split at random between the two sets, keeping the number of negative and positive examples equal within each set.
Table \ref{table:data_sets} indicates 
the statistics of the resulting  
datasets,
which will be released as part of this work.

\subsection{Manually labelled dataset - SentiFM}
To the best of our knowledge, the only manually annotated dataset for event detection in news articles is SentiFM \cite{jacobs2018economic}. This dataset contains manual annotations of sentences into $10$ predefined financial event types. However, this dataset is designed to solve a slightly different problem from the one explored in this paper. SentiFM was constructed in the context of a multi-class classification problem, whereas here we deal with a binary problem. Namely, we are not interested in event types, and do not assume there is a closed set of underlying types describing the events of interest. Indeed, it is possible that an event of interest might not be included in the SentiFM taxonomy, and hence a corresponding sentence would be labeled as negative.
Despite these differences, we sought to examine how a classifier trained on the SentiFM data would perform on our task. 
To this end, we created a \textit{binary} version of SentiFM, by considering all 'no-event' sentences as negative examples, and all event types as positives. We kept the original train/test split (see Table \ref{table:data_sets}) and denote this data set as $SentiFM$-$binary$.

\subsection{2019 News Sentences - $News$-$2019$}
\label{eval_data}
In order to evaluate methods for detecting company-related events within news data, we compile a set of sentences from news articles. Specifically, we selected the $10$ S\&P companies with the largest number of events from 2019 mentioned in their Wikipedia page (see Table \ref{table:news_eval_data}). For each company, we retrieved all articles from 2019 on Seeking Alpha\footnote{seekingalpha.com; Transcriptions of company earning calls were filtered out due to their unique nature.} that contained the company name in their title. We assume that this set of articles provides a good coverage of the company's events of interest during 2019. We applied  sentence-splitting\footnote{using the NLTK library} 
on the retrieved articles, keeping only sentences 10-50 tokens long. 
\begin{table}
\begin{tabular}{|l|c|c|} 
\hline
Model & Train & Test \\
\hline
\small
$SentiFM$-$binary$ & 8943 (0.2) & 443  (0.2) \\
$S\&P$-$wiki$ &  6130 (0.5) & 272 (0.5)	 \\
$Extended$-$wiki$ & 20074 (0.5) & 908 (0.5)	 \\

\hline
\end{tabular}
\caption{Data size (number of sentences) for the three models. The numbers in parenthesis indicate the percentage of positive samples.}
\label{table:data_sets}
\end{table}

\begin{table}
\begin{center}
\begin{tabular}{|l|c|c |} 
\hline
Company	& Articles	& Sentences \\
\hline
\small Apple Inc.	& \small 438	& \small 10627 \\
\small Facebook	& \small 302	& \small 6827 \\
\small Qualcomm	& \small 120	& \small 3332 \\
\small FedEx		& \small 67	& \small 1808 \\
\small Anadarko Petroleum	& \small 91	& \small 1478 \\
\small Xilinx		& \small 53	& \small 569 \\
\small MGM Resorts International	& \small 32	& \small 463 \\
\small Accenture	& \small 24	& \small 442 \\
\small Allergan	& \small 35	& \small 421 \\
\small Campbell Soup Company	& \small 27	& \small 307 \\
\hline
\end{tabular}
\end{center}
\caption{Number of articles and sentences in the $News$-$2019$ evaluation data.}
\label{table:news_eval_data}
\end{table}

\section{Experiments}
\label{experiments}
The datasets described in Section \ref{data} were used to train three event detection models.
All classification models are based on BERT \cite{devlin2018bert}, which has shown state-of-the-art results in many NLP tasks. We use a single-sentence input, and fine-tune the classifier with the $SentiFM$-$binary$, $S\&P$-$wiki$ and $Extended$-$wiki$ data sets. Henceforth, we will use these names to refer to their corresponding BERT models. We use the BERT\textsubscript{BASE} model configuration, with maximum sequence length of 256, batch size of 16, dropout rate of 0.1 and learning rate of 5e-5. Each model was fine-tuned over 3 epochs, using a cross-entropy loss function.

\subsection{Initial model evaluation}
We first evaluate the performance of the three models on their corresponding test sets. As shown in Table \ref{table:res_on_dev}, all models reach high performance when tested on the same type of data used in training.
Next, we evaluate these models on the $Extended$-$wiki$ test set (see Table \ref{table:res_on_dev}). Notably, although less than $15\%$ of the companies in $Extended$-$wiki$ are in $S\&P$-$wiki$, the latter model exceeds $90\%$ precision and recall over the $Extended$-$wiki$ test data. This suggests 
that the model is also able to detect events for companies that were not seen in training.   
 
\subsection{Identifying Wikipedia events in the news}
Ultimately we are interested in the ability to detect events in the target domain of \textit{news articles}.
To validate performance over this domain, we used sentences from $News$-$2019$ and cross-referenced them with company events from Wikipedia. Specifically, we manually extracted events from 2019 from the Wikipedia pages of the companies in Table \ref{table:news_eval_data}. For each event, we asked 3 annotators to mark all sentences from $News$-$2019$ which mention this event. In total, 26 of the Wikipedia events were mentioned in at least one sentence.

We then applied each of the three models to all the news sentences, and kept only the sentences that were classified as positive by the model.
For each model, we measure the event recall rate as the fraction of Wikipedia events which are mentioned in at least one positively-classified sentence.

As expected, the recall rates of the Wikipedia-based models over the news data (Table \ref{table:wiki_events_in_news}) are lower than those achieved over Wikipedia data. This may be due to the difference in writing style between the two sources.   
Notably, even though $SentiFM$-$binary$ was trained on news data, its recall is the lowest among the three models. This may be attributed to the mismatch between the event types in $SentiFM$ and those in Wikipedia.

Sorting the positively-classified sentences by their model score, we also measure the average rank of the highest-scored mention of each event (Table \ref{table:wiki_events_in_news}).
Clearly, the Wikipedia reference events do not fully cover all company-related events that occurred over this time period. Still, since we presume events mentioned in Wikipedia are relatively significant, we expect a good event-detection model to rank them among its top predictions.

\begin{table}
\begin{tabular}{|l|c|c|c|c |} 
\hline
Model	& Precision & Recall & F1 \\
\hline
\hline
$SentiFM$-$binary$ & 0.97 & 0.96 & 0.96 \\
$S\&P$-$wiki$ & 0.97 & 0.92 & 0.94 \\
$Extended$-$wiki$  & 0.93 & 0.95 & 0.94 \\
\hline
\hline
$SentiFM$-$binary$ & 0.80 & 0.30 & 0.44 \\
$S\&P$-$wiki$ & 0.92 & 0.93 & 0.93 \\
$Extended$-$wiki$  & 0.93 & 0.95 & 0.94 \\
\hline
\end{tabular}
\caption{Model performance on its test set (upper) and on the $Extended$-$wiki$ test set (lower)}
\label{table:res_on_dev}
\end{table}

\begin{table}
\begin{center}
\begin{tabular}{|l|c|c|}
\hline
Model & Recall & Avg. Rank \\
\hline
$SentiFM$-$binary$ & 0.38 & 153 \\
$S\&P$-$wiki$ &  0.73 & 21 \\
$Extended$-$wiki$  & 0.77 & 19 \\
\hline
\end{tabular}
\end{center}
\caption{Model performance for identifying 26 Wikipedia events in the news data.}
\label{table:wiki_events_in_news}
\end{table}

\subsection{Identifying general events in the news}
So far our experiments considered only Wikipedia events. However, there are likely numerous company-related news events that are not necessarily mentioned in the company's Wikipedia page. Thus, the question remains whether the Wikipedia-based models are able to detect such events as well. To this end, the top 20 model predictions of $SentiFM$-$binary$ and $Extended$-$wiki$ for the companies in Table \ref{table:news_eval_data} were annotated
by three co-authors of this work. 
The guidelines were to
determine whether a given sentence contains information which may have influence on the company’s stock price, 
as such events presumably deserve the attention of a finance analyst.  
The annotation process was composed of two stages. First, each sentence was annotated by two labelers. Then, the sentences on which there was disagreement between the labelers ($21\%$ of the sentences) were annotated by a third annotator. Average agreement between the initial two annotators was $0.45$ (Cohen's Kappa).

Table \ref{table:top20_events_in_news} shows the precision of the two models, compared to a baseline of 
randomly-selected sentences. The $Extended$-$wiki$ model outperforms $SentiFM$-$binary$.

\begin{figure}[ht!]
\centering
\includegraphics[width=75mm]{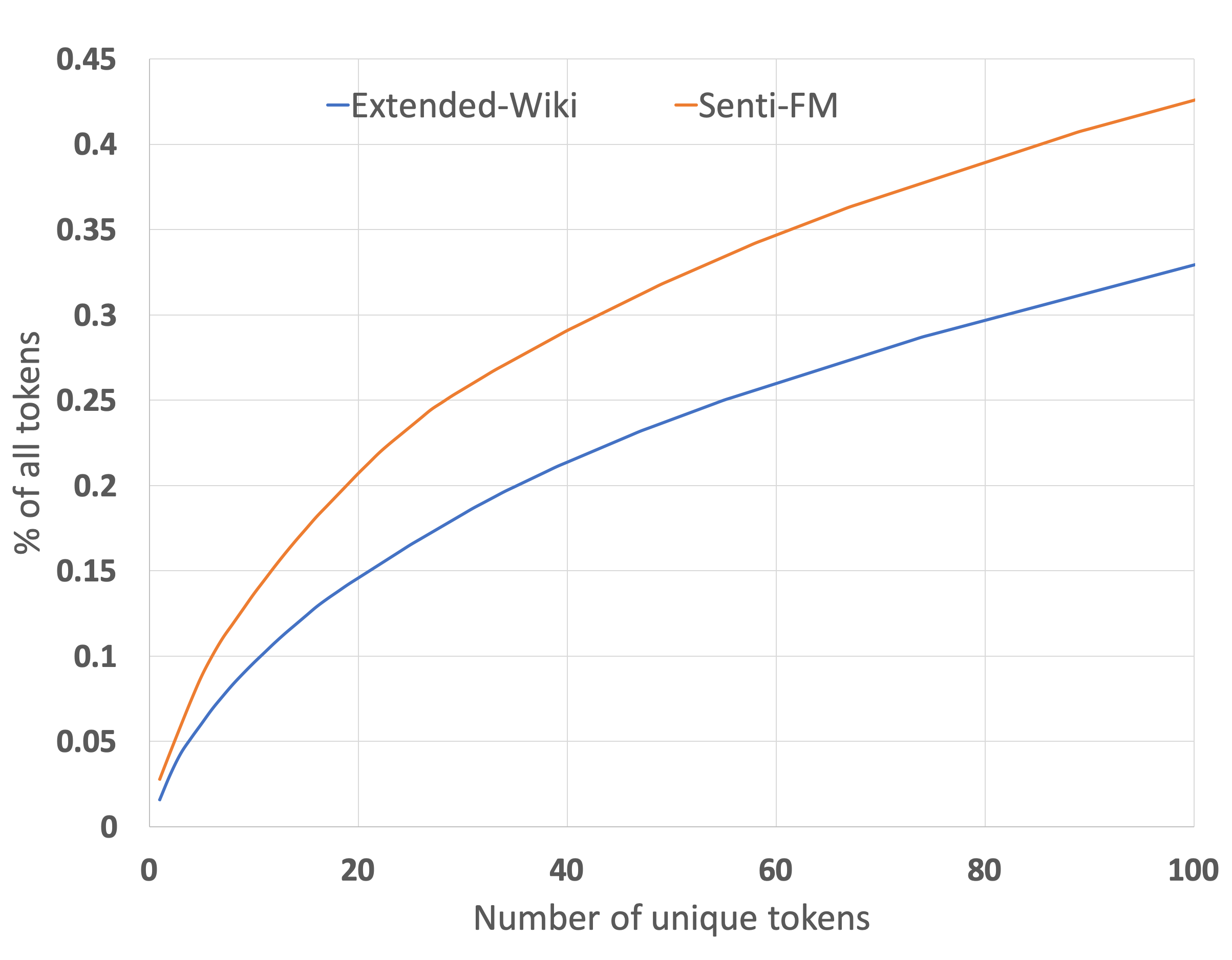}
\caption{Cumulative token frequency over top model predictions. \label{fig:tokens}}
\end{figure}

Finally, we wanted to analyze the diversity of events captured by the two models. For this purpose, we looked at the distribution of unique tokens in the top 200 predictions of each model, after filtering out stop words and the companies appearing in the list of Table \ref{table:news_eval_data}. We sorted the remaining tokens by their frequency from highest to lowest, and computed the cumulative frequency as a function of the number of unique tokens. Figure \ref{fig:tokens} indicates that the top candidates of $Extended$-$wiki$ capture a richer vocabulary than $SentiFM$-$binary$, which is dominated by a smaller group of tokens. 
For example, $20\%$ of the tokens are covered by the 36 and 19 most frequent tokens in $Extended$-$wiki$ and $SentiFM$-$binary$, respectively. 
Moreover, despite their similar precision values, the population of events captured by the two models is quite different - the overlap between their top candidates is less than $10\%$ (18 out of 200 examples). This observation suggests that the models are complementary, and that there is potential benefit to combining them.

\begin{table}
\begin{center}
\begin{tabular}{|l|c|}
\hline
Model & Precision \\
\hline
$Random$ $sentences$ & 0.28\\
$SentiFM$-$binary$ & 0.70 \\
$Extended$-$wiki$  & 0.74 \\
\hline
\end{tabular}
\end{center}
\caption{Average precision over the top-20 predicted events in the news evaluation data.}
\label{table:top20_events_in_news}
\end{table}

\section{Discussion}
This paper 
focused on detecting 'important'  
events in news articles,
related to a specific company. 
We suggested to leverage information contained in Wikipedia to create weakly-labelled data, and proved the usefulness of the resultant classifier for the desired task.
We believe that the results can be further improved by finding additional sources for weak-labels, 
e.g. by exploiting information from relevant knowledge bases. 

 The potential coverage of relevant events can be increased by retrieving articles which do not necessarily include the name of the considered company in their title. 
Extending our framework to pinpoint noteworthy events for a particular company, mentioned in articles that are not focused on that company, is a natural direction for future research.
Such an extension will require adapting the weak labelled data and the corresponding classifiers to cope with an environment in which sentences are not necessarily relevant to the company. 

\bibliography{main}
\bibliographystyle{acl_natbib}
\newpage 
\appendix

\end{document}